# Online Sparse Streaming Feature Selection with Uncertainty

Feilong Chen
*School of Computer Science and Technology,*
*Chongqing University of Posts and Telecommunications*
*Chongqing Institute of Green and Intelligent Technology, Chinese Academy of Sciences*
*Chongqing School, University of Chinese Academy of Sciences*
Chongqing 400714, China
s200231207@stu.cqupt.edu.cn

Di Wu
*College of Computer and Information Science,*
*Southwest University,*
Chongqing 400715, China,
wudi.cigit@gmail.com

Jie Yang
*School of Physics and Electronic Science, Zunyi Normal University,*
Zunyi 563002, China
yj530966074@foxmail.com

Yi He
*Department of Computer Science, Old Dominion University,*
Norfolk, VA 23529, USA
yihe@cs.odu.edu

*Abstract*—Online streaming feature selection (OSFS), which conducts feature selection in an online manner, plays an important role in dealing with high-dimensional data. In many real applications such as intelligent healthcare platform, streaming feature always has some missing data, which raises a crucial challenge in conducting OSFS, i.e., how to establish the uncertain relationship between sparse streaming features and labels. Unfortunately, existing OSFS algorithms never consider such uncertain relationship. To fill this gap, we in this paper propose an online sparse streaming feature selection with uncertainty (OS$^2$FSU) algorithm. OS$^2$FSU consists of two main parts: 1) latent factor analysis is utilized to pre-estimate the missing data in sparse streaming features before conducting feature selection, and 2) fuzzy logic and neighborhood rough set are employed to alleviate the uncertainty between estimated streaming features and labels during conducting feature selection. In the experiments, OS$^2$FSU is compared with five state-of-the-art OSFS algorithms on six real datasets. The results demonstrate that OS$^2$FSU outperforms its competitors when missing data are encountered in OSFS.

## I. Introduction

Feature selection is an important technique to reduce high-dimensional data [1]. It selects the best subset from all high-dimensional features as well as retains their essential characteristics. Traditional feature selection methods, such as mutual information-based [1], [2], neighborhood rough set-based [3]–[5], mRMR (minimal Redundancy and Maximal Relevance)-based [6], assume that the feature space is predefined and static. But in many real applications, it is impossible to observe all the feature space in advance as features usually generate over time continually. Hence, the assumption of static feature space does not hold in real-world scenarios.

Considering dynamic feature space, online streaming feature selection (OSFS) methods were proposed [7]. OSFS performs feature selection on streaming features that arrive one by one over time with instance count remaining fixed. Recently, many OSFS algorithms were proposed, including fast-OSFS [7], SAOLA [8], OSFASW [9], SFS_FI [10], etc. Notably, these OSFS algorithms have a common assumption that streaming features are complete without any missing data. While in real applications, it is impossible to collect dynamic features completely [11]. For example, the features describing patients' symptoms involve many different healthcare service providers (hospitals, labs, etc.) and inspection equipment (respiratory sensors, pulse monitors, *etc*.). Since different patients are prone to have different symptoms, it is impossible that each patient can collect all the features. Therefore, a new problem arises: how to implement online sparse streaming feature selection (OS$^2$FS)?

To address this problem, Wu et al. [12] proposed a latent-factor-analysis-based online sparse-streaming-feature selection algorithm (LOSSA). LOSSA adopts latent factor analysis (LFA) [13]–[23] to pre-estimate the missing data of sparse streaming features. However, since there are some inevitable errors between estimated data and real data, it is difficult to establish a certain relationship between sparse streaming features and labels. As a result, uncertainty pervades the process of OS$^2$FS. Unfortunately, LOSSA never considers such uncertainty, which may cause inferior performance.

To fill this gap, we proposed an <u>o</u>nline <u>s</u>parse <u>s</u>treaming <u>f</u>eature <u>s</u>election with <u>u</u>ncertainty (OS$^2$FSU) algorithm. Significantly different from LOSSA, OS$^2$FSU employs fuzzy logic and neighborhood rough set to alleviate the uncertainty between sparse streaming features and labels during implementing OS$^2$FS, making it achieve better performance. The main contributions of this paper are summarized as follows:

a) Algorithm. We propose OS$^2$FSU to address the two problems of missing data and uncertainty in OS$^2$FS.

b) Theoretical analyses. We provide detailed technical principles, algorithm design, and time complexity analysis for the proposed OS$^2$FSU.

c) Experiments. We compare the proposed OS$^2$FSU with five algorithms on six real datasets. The results demonstrate that OS$^2$FSU outperforms these comparison algorithms when missing data are encountered in OSFS.

## II. RELATED WORK

OSFS is an important and hot research direction of feature selection. To date, many sophisticated OSFS algorithms were proposed [7], [24], [25].

A classical OSFS framework was proposed by Wu *et al.* [7]. They proposed to use the online relevance and online redundancy analyses to detect the irrelevant and redundant features during OSFS. Yu et al. [8] proposed SAOLA algorithm for large-scale OSFS. SAOLA employs a novel pairwise comparison technique to maintain a frugal model over training time. To reduce the number of selected features, OSFASW was proposed by You *et al* [9] based on a self-adaptive sliding-window sampling strategy. Later, Zhou et al. proposed OFS-Density [24] and OFS-A3M [25] algorithms based on the neighborhood rough set. Both of them do not consider domain information during training. OFS-Density uses the density information of the neighborhood instances, while OFS-A3M uses the gap information between neighbors of the target object, to select the proper final feature set automatically. Consequently, considering the feature interaction, SFS-FI [10] measures the interaction degree between features by interaction gain to guarantee that the selected features interact with each other. Notably, the above OSFS algorithms do not consider the missing data issue of OSFS. To address this problem, Wu et al. [12] proposed a LOSSA algorithm, adopts latent factor analysis (LFA) to pre-estimate the missing data of sparse streaming features.

Since there are some inevitable errors between estimated data and real data, uncertainty pervades the process of OSFS, which is not considered by LOSSA. Significantly different from LOSSA, OS$^2$FSU employs fuzzy logic and neighborhood rough set to alleviate the adverse effects of such uncertainty, making it achieve much better performance than LOSSA.

## III. PRELIMINARIES

### A. Latent Factor Analysis (LFA)

LFA[13]–[16], [19], [20], [26]–[41] is a decomposition-based matrix method that specializes in the high-dimensional and sparse matrix, aims to get a complement matrix by factorizing the observed elements into two low-rank matrices.

**Definition 1 (Latent Factor Analysis).** For a given sparse matrix $R^{M \times J}$, LFA factorizes into two low-rank matrices $P^{M \times d}$ and $Q^{J \times d}$. If the value of predicted matrix $\hat{R}$' is similar to $R$ at the known position, then it can be considered that the value is also approximately in the predicted position, where $\hat{R} = P \times Q^T$. Loss function (1) used to measure the sum errors between $\hat{R}$ and $R$, and add regularization to avoid overfitting.

$$L(P,Q) = \frac{1}{2}\left\|\Omega \odot \left(R - PQ^T\right)\right\|_F^2 + \frac{\lambda}{2}\left(\|P\|_F^2 + \|Q\|_F^2\right) \quad (1)$$

Where $\lambda$ is the regularization parameter, $\odot$ means Hadamard product and $\Omega_{m,j}$ equals 1 if the data $r_{m,j}$ at the $m$-th row and $j$-th column is observed, otherwise equals 0. $P$ and $Q$ can be acquired conveniently by stochastic gradient descent(SGD).

### B. Fuzzy Logic

Fuzzy logic[42]–[44] is a fuzzy method to deal with uncertain problems which can't be solved by boolean logic. Whether the elements in the fuzzy set are discrete or continuous, the membership function(MF)[45] portrayed the ambiguity, and usually used to handle some part of model uncertainties. The common MFs introduced in follows.

1) Triangular MF: the boundary varies linearly from 0 to 1 membership grade, and only one discrete element can take membership degree as 1. It can be described by the following formula:

$$\mu_A(x) = \begin{cases} 0, & x \leq a \\ (x-a)/(b-a), & a < x \leq b \\ (x-c)/(b-c), & b < x < c \\ 0, & x \geq c \end{cases} \quad (2)$$

2) Trapezoidal MF: it is similar to triangular MF, but all elements in a certain range can obtain the maximum membership. It can be characterized by the following formula:

$$\mu_A(x) = \begin{cases} 0, & x < a \\ (x-a)/(b-a), & a < x \leq b \\ 1 & b < x < c \\ (d-x)/(d-c), & c \leq x \leq d \\ 0, & x \geq d \end{cases} \quad (3)$$

Gaussian MF: the boundary varies nonlinearly from the lowest to the highest membership degree. Let σ denote the standard deviation and use curve-fitting constant $c$ to curve-shape control. Gaussian MF is represented by:

$$\mu_A(x) = e^{-(x-c)^2/\sqrt{(2\sigma)}} \quad (4)$$

The common MFs graphs shows in Figure.1.

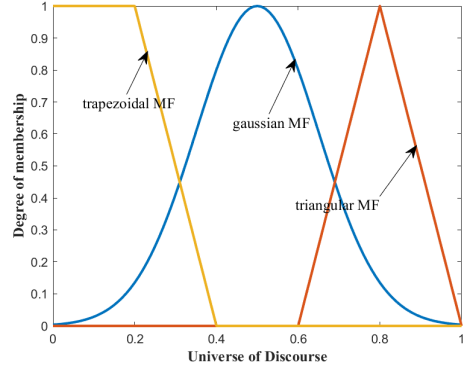

Fig. 1. Graphic of fuzzy membership function

*C. Neighborhood Rough Set*

For a neighbor rough set $NDS = <U,F,D>$[3]–[5], where $U = \{x_1, x_2,...,x_n\}$ represents a finite set of $n$ samples, $F$ is the set of features. $D$ represents the decision attribute. Then, we have the following definitions for $NDS$.

**Definition 2.** Given a nonempty metric space $<U, F, D>$, let $x \in U$, $h \geq 0$, $G \subseteq F$, $X \subseteq U$, so the point set:
$$\delta_G^h(x) = \{y \mid \Delta(x, y) \leqslant h, y \in U\} \quad (5)$$
denotes the neighborhood of arbitrary $x$ on attribute subset $G$, centered at $x$ and $h$ is the distance radius. So the lower approximation and upper approximation of decision $D$ on $G$, in terms of a $\delta$ neighborhood relation are defined, respectively:
$$\underline{R_G^\delta} X = \{x_i \mid \delta_G(x_i) \subseteq X, x_i \in U\} \\ \overline{R_G^\delta} X = \{x_i \mid \delta_G(x_i) \cap X \neq \emptyset, x_i \in U\} \quad (6)$$

**Definition 3**: Given $NDS=<U,F,D>$, $G \subseteq F$, the maximum nearest neighbor dependency degree of $G$ on the decision attribute $D$ is given as follows:
$$\gamma_G(D) = \frac{Card(\underline{R_G}D)}{Card(U)} \quad (7)$$

## IV. OUR PROPOSED ALGORITHM

*A. Problem statement*

**Definition 4 ( *Sparse Streaming features*[12]).** Let sparse streaming features set $F=\{F_1, F_2, …, F_T\}$ where the feature vector $F_t=[f_{1,t}, f_{2,t}, …, NAN, …, f_{M,t}]^T$, $t \in \{1, 2, …, T\}$. $F$ is a feature set that has $M$ instances and $T$ features with missing data. $F_t$ means a sparse feature flowed at time point $t$, $NAN$ denotes missing data.

**Definition 5 (*Online Spares Streaming Feature Selection, OS²FS*[12]).** Let $O_{t-1} = \{F_1, F_2, …, F_{t-1}\}$ denotes the observed streaming features set arrived at time point $t$-1. The selected spares streaming feature subset $S_{t-1} \subseteq O_{t-1}$. At time point $t$, a new sparse feature $F_t$ arrives, OS²FS accesses the feature relevance to label $C$, and decides add it to $S_{t-1}$ or discard it, ensuring the model with maximum performance at time point $t$.

For OS²FS, there is exist two challenges: 1) completing the missing data of the sparse streaming features, and 2) establishing the uncertain relationships between sparse streaming features and labels caused by complement.

*B. Our Algorithm*

Our proposed algorithm OS²FSU can be divided into two phases to handle the above two challenges. The first part is complete the missing value of the sparse streaming features and the second part is to select the best-classified prediction subset with fuzzy logic. The framework of our OS²FSU is shown in Fig.2.

**Phase I:** For sparse streaming features, LFA predicts the missing elements based on the observed data. Let $B^{M \times Bs}$ denote the buffer matrix that received the streaming features. As the features flowed from time point $t$ to $t+B_S-1$, i.e., $B = \{F_t, F_{t+1}, …, F_{t+Bs-1}\}$. After $B$ is filled, LFA will be applied to complete the missing value in $B$, and let $\hat{B} = \{\hat{F}_t, \hat{F}_{t+1}, …, \hat{F}_{t+Bs-1}\}$ denotes the corresponding completed matrix. According to definition 1, we reformulate the objective function for every single entry in $B$ as follows:
$$\varepsilon_{m,j} = \frac{1}{2}\left(f_{m,j} - \sum_{v=1}^{d} p_{m,v} q_{j,v}\right)^2 + \frac{\lambda}{2}\left(\sum_{v=1}^{d} p_{m,v}^2 + \sum_{v=1}^{d} q_{j,v}^2\right). \quad (8)$$

Where $d$ is the rank of $B$, $f_{m,j}$ is the observed element of $m$-th row of $F_j$. As SGD has an excellent iterative effect, we minimize $\varepsilon_{m,j}$ by SGD. In one iteration, all observed entries in B are used to train $P$ and $Q$ unceasingly. Then $\hat{B} = PQ^T$.

**Phase II:** This phase conducts feature selection for all the features in $\hat{B}=\{\hat{F}_t, \hat{F}_{t+1}, …, \hat{F}_{t+Bs-1}\}$, and contains three main operations, i.e., online relevance analysis, online redundancy analysis and online fuzzy correlation analysis.

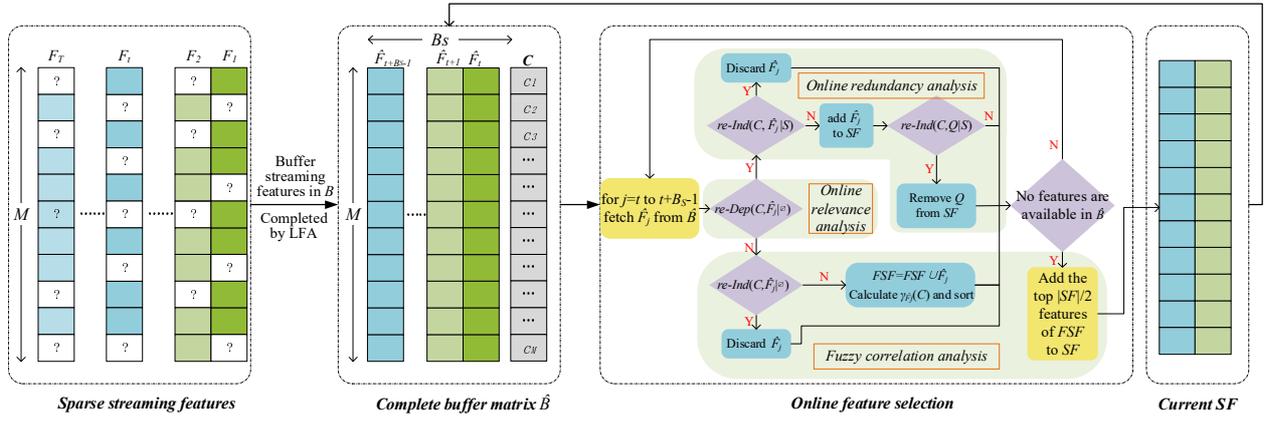

Fig. 2. The framework of OS²FSU algorithms

TABLE I. Algorithm Design of OS²FSU

| | |
|---|---|
| | **Input:** sparse streaming features $F = \{F_1, F_2, \ldots, F_T\}$. |
| 1 | Parameters: maximum number of iteration *Imax*, the number of |
| 2 | latent factor *d*, the fixed column count $B_S$ |
| 3 | Initialize: $SF=\emptyset$, $FSF=\emptyset$, $B=\emptyset$, $B_S$ |
| 4 | **Repeat** |
| 5 |   A new sparse streaming feature $f_t$ arrives at the time $t$; |
| 6 |   **if** $|B|\neq B_S$ |
| 7 |     $B = B \cup F_t$ |
| 8 |     **continue** |
| 9 |   **initializing** $d, \lambda, \eta, P, Q$, *Imax*, *iter*=1 |
| 10 |   **While** *iter* < *Imax* && not converge |
| 11 |     **for** $j=t$ to $t+B_S-1$ |
| 12 |       **for** each $f_{m,j} \in K_j$ // $K_j$ is the known entry set of $F_t$ |
| 13 |         update $p_m, q_j$ according to (1) |
| 14 |       **end for** |
| 15 |     **end for** |
| 16 |     *iter* = *iter*+1 |
| 17 |   **end while** |
| 18 |   $\hat{B}=PQ^T$ |
| 19 |   /\***online relevance analysis**\*/ |
| 20 |   **for** $i = t$ to $t+B_S-1$ |
| 21 |     fetch $\hat{F}_j$ from $\hat{B}$ |
| 22 |     **If** $re\text{-}Dep(C, \hat{F}_j|\emptyset)$ |
| 23 |       /\***online redundancy analysis I**\*/ |
| 24 |       **If** $\exists S \subseteq SF$ s.t. $re\text{-}Ind(C, \hat{F}_j \mid S)$ |
| 25 |         discard $\hat{F}_j$ and continue |
| 26 |       **end if** |
| 27 |       $SF = SF \cup \hat{F}_j$ |
| 28 |       /\***online redundancy analysis II**\*/ |
| 29 |       **for** each feature $Q \in SF\text{-}\hat{F}_j$ |
| 30 |         **if** $\exists S \subseteq SF\text{-}Q$ s.t. $re\text{-}Ind(C, Q|S)$ |
| 31 |           $SF = SF - Q$ |
| 32 |         **end if** |
| 33 |       **end for** |
| 34 |     **else if** not $re\text{-}Ind(C, \hat{F}_j \mid \emptyset)$ |
| 35 |       /\***online fuzzy correlation analysis** \*/ |
| 36 |       $FSF = FSF \cup \hat{F}_j$ |
| 37 |       Calculated $\gamma_{\hat{F}_j}(C)$ by (7) and sorted in $FSF$ |
| 38 |     **end if** |
| 39 |   **end for** |
| 40 |   Add the top $|SF|/2$ of $FSF$ to $SF$ |
| 41 |   $B=\emptyset$, $t=t+1$ |
| 42 | **Until** no features are available |
| | **Output** selected completed streaming features subset $SF$ |

**Online relevance analysis:** At time point $t$, we fetch $\hat{F}_j$ from $\hat{B}$ and evaluate its relevance to label $C$. For continuous data, *Fisher's-z test* [7] is used to calculate the relevance. Let $p$ denotes the $p$-value return by *Fisher's-z test*. Significance level $\alpha$ is used to measure the probability of rejecting the null hypothesis, if $p>\alpha$, the null hypothesis is accepted, $\hat{F}_j$ and $C$ are conditionally independent within given $S$, called $Ind(C, \hat{F}_j \mid S)$. if $p \leq \alpha$, the null hypothesis is rejected, defined $Dep(C, \hat{F}_j \mid S)$ to describe the conditionally dependent between $\hat{F}_j$ and $C$ within given $S$. For discrete data, $G^2$ test will be implemented.

**Online redundancy analysis:** If $\hat{F}_j$ is relevant to label, then we analyze its redundancy. If $\exists S \subseteq MB(C)_{t-1}$, s.t. $P(C | \hat{F}_t, S) = P(C | S)$, $\hat{F}_j$ is the redundant feature and discarded. Furthermore, we also evaluate the features exited in $MB(C)_{t-1}$ if become redundant as $\hat{F}_j$ added, where $MB(C)$ is the Markov blanket for $C$.

**Fuzzy correlation analysis:** Traditionally, significance level $\alpha$ is 0.01 or 0.05. But in OS²FS, due to the uncertain error between real value and predicted value, a fixed $\alpha$ is no longer acceptable. So our algorithm utilizes trapezoidal MF to regulate $\alpha$ fluctuating between [0.01,0.1], then we define the fuzzy relevance feature and redefine $Ind(C, \hat{F}_j | S)$ and $Dep(C, \hat{F}_j | S)$.

***Definition 8*** (*Fuzzy Relevance Feature*). Let $\mu$ denote fuzzy significant threshold. The *p*-value of feature $\hat{F}_j$ is calculated by relevance analysis, if $\mu < p \leqslant 0.1$, we call $\hat{F}_j$ is fuzzy relevance feature. And will be calculated its dependency degree $\gamma_{\hat{F}_j}(C)$ by (7) and sorted.

***Definition 9***: If $p > 0.1$, we accept the null hypothesis that $\hat{F}_j$ and label are conditionally independent within given $S$, called re-$Ind(C, \hat{F}_j | S)$. If $p \leqslant \mu$, reject the null hypothesis, $\hat{F}_j$ is the correlation with $C$ within given $S$, defined re-$Dep(C, \hat{F}_j | S)$.

**Algorithm Design.** Based on the theorem analysis, the pseudocode of OS²FSU is given in Table I. Step 5-8 uses buffer matrix $B$ to cache the received features. When $B$ is filled, steps 9-18 implement LFA model to predict the missing data, and get the completed matrix $\hat{B}$. For each feature $\hat{F}_j$ in $\hat{B}$, Steps 19-22 utilize re-$Dep(C, \hat{F}_j|\emptyset)$ to evaluate the correlation between $\hat{F}_j$ and $C$. If $\hat{F}_j$ is relevant, it will go to step 23 to online redundancy analysis. In steps 23-34, redundancy analysis I evaluate the redundancy of the new relevant feature $\hat{F}_j$. If $\hat{F}_j$ is redundant, it will be discarded. Otherwise, $\hat{F}_j$ will be added to the currently selected feature set $SF$, then redundancy analysis II is triggered. The original feature exiting in $SF$ will be checked to whether become redundant as the new feature $\hat{F}_j$ added. If $\hat{F}_j$ is neither relevance feature nor irrelevance feature, the dependency degree $\gamma$ of $\hat{F}_j$ will be calculated and sorted in steps 35-40, the top $|SF|/2$ of $FSF$ will be added into $SF$.

*C. Time complexity Analysis.*

The time complexity of our algorithm mainly depends on the missing data completion. Supposing there are $M$ instances and $T$ features, the rate of total missing is $\theta$, the time complexity of Phase I is $O(M \times T \times (1-\theta) \times d)$. For Phase II, let $RF$ denotes all relevant features in the total feature set, $RF_1$ is the performed redundant features in $RF$. The time complexity is $O(|RF-RF_1|\omega^{|SF|} + |RF_1||SF|\omega^{|SF|})$, where $\omega$ considers all the subsets of $SF$. Thus the total time cost is $O(M \times T \times (1-\theta) \times d + |RF-RF_1|\omega^{|SF|} + |RF_1||SF|\omega^{|SF|})$.

## V. EXPERIMENTS AND RESULTS

*A. Experimental Settings*

1) **Dataset.** We apply our algorithm on six datasets from ASU, openML, UCI. They are summarized in Table II.

TABLE II. EXPERIMENTAL DATASETS.

| Mark | Datasets | Features | Instances | Classes |
|---|---|---|---|---|
| D1 | mfeat-factors | 217 | 2000 | 2 |
| D2 | SMK-CAN-187 | 19993 | 187 | 2 |
| D3 | lung | 3312 | 203 | 4 |
| D4 | USPS | 256 | 9298 | 10 |
| D5 | COIL | 241 | 1500 | 6 |
| D6 | Isolet | 617 | 1560 | 26 |

2) **Baselines.** To validate our algorithm is excellent, we compared OS²FSU with five algorithms, i.e., LOSSA[12], fast-OSFS[7], SAOLA[8], OSFASW[9], and SFS_FI[10].

3) **Experimental Designs.** For OS²FSU and LOSSA, we default set $B_S$=15, $\lambda$=0.01, and $\eta$=0.00001. Note that fast-OSFS, SAOLA, OSFASW, and SFS_FI are used to deal with completed dataset, so we fill the missing data by zero or means. If filled zero we renamed by OSFS_Z, SAOLA_Z, OSFASW_Z, SFS_FI_Z. If filled means, we renamed by OSFS_M, SAOLA_M, OSFASW_M, SFS_FI_M.

To evaluate the classification accuracy of the selected features, we employ three classifiers KNN, SVM, and Random Forest, then take their average value as the comparison result. To obtain objective results, 5-fold cross-validation was performed. Note the testing data was not sparse, while the training data have different missing rates from 0.1 to 0.9. We conduct all experiments on a PC with Intel(R) i7-9700 3.00GHz CPU, and 16G RAM.

*B. Results and Discussions of Comparison Experiments.*

1) **Details of selected features.** We evaluate the number of selected features of all the algorithms on each dataset at $\theta$=0.1. Table III presents the comparison results, where we see that OS²FSU tends to select a moderate number of features.

2) **Comparison results on all datasets with $\theta$=0.1.** Focus on $\theta$ = 0.1, the detail results of ten algorithms execution are recorded in Table IV. OS²FSU outperforms other algorithms in 6/6 datasets. Comparing the average classification accuracy of six datasets, OS²FSU is 10.29% higher than the ranked second algorithm LOSSA, and beyond 19.81% than the algorithm SAOLA_M which ranked last.

*3) Comparison results on D1 with different θ*. We evaluate the classification accuracy on D1 when $\theta$ increases from 0.1 to 0.9. Table V shows the detail average classification accuracy of three classifiers comparison results of OS²FSU with other algorithms on D1. On all test cases, OS²FSU achieves better classification accuracy than the other algorithms.

*4) Comparison results on all datasets with different θ*. We evaluate each algorithm's average accuracy when $\theta$ increases from 0.1 to 0.9 on each dataset and present in Fig 3. We observe that OS²FSU achieves higher accuracy in most cases. And when $\theta$ is smaller than 0.6, the classification accuracy does not evidently decrease or even has a slight improvement, while when $\theta$ is greater than 0.6, the classification accuracy decreases rapidly, which maybe because the errors between estimated and real data become significant.

*5) **Statistical analysis***. To effectively explore the statistical significance of our algorithms, the Wilcoxon signed-ranks test[46] will be used. Based on each algorithm's average accuracy of three classifiers, we compare the six results of OS²FSU with that of other algorithms one by one by the Wilcoxon signed-ranks. In the statistical test, we sort the differences of pairwise samples in sequence to obtain the six ranks $\{R_1, R_2, ..., R_6\}$, where the rank value of the smallest difference is 1 and that of the largest difference. Then, the two indicators of positive-rank sum $R^+$ and negative rank-sum $R^-$ are computed as follows:

$$R^+ = \sum_{1 \leq i \leq 6, R_i > 0} R_i, \quad R^- = \sum_{1 \leq i \leq 6, R_i < 0} R_i, \quad (9)$$

where $R^+$ and $R^-$ denote the rank sum of OS²FSU and each comparison algorithm, respectively. Then, let $R_m = min(R^+, R^-)$, the statistics $z$ is calculated by:

$$z = \frac{R_m - \frac{1}{4}N(N+1)}{\sqrt{\frac{1}{24}N(N+1)(2N+1)}}, \quad (10)$$

where $N$ denotes the number of datasets. According to the standard normal distribution table, the null-hypothesis will be rejected if $z$ is smaller than -1.64 with a significance level $\alpha = 0.1$. So when $N=6$, the null-hypothesis will be rejected if $Rm$ is less than or equal 2.

TABLE III. THE SELECTED FEATURE NUMBER ON $\theta=0.1$

| Algorithms | D1 | D2 | D3 | D4 | D5 | D6 | Average |
|---|---|---|---|---|---|---|---|
| OS²FSU | 13.2 | 13.4 | 20.6 | 17.2 | 9.6 | 26.4 | 16.73 |
| LOSSA | 7.6 | 5.2 | 9.8 | 10.8 | 6.2 | 14.2 | 8.97 |
| SFS_FI_Z | 4 | 5.2 | 34.6 | 12.6 | 5 | 7.6 | 11.50 |
| OSFS_Z | 13.6 | 5.2 | 9.4 | 17.6 | 13.8 | 18.2 | 12.97 |
| SAOLA_Z | 6.6 | 5.2 | 66 | 5.2 | 4.8 | 6.8 | 15.77 |
| OSFASW_Z | 13 | 5.2 | 8.6 | 16 | 13.6 | 15.6 | 12.00 |
| SFS_FI_M | 4.8 | 5.2 | 15.4 | 10.4 | 4 | 8.4 | 8.03 |
| OSFS_M | 7.8 | 5.2 | 11.4 | 10.2 | 6.8 | 14.2 | 9.27 |
| SAOLA_M | 5 | 5.2 | 33.2 | 4.8 | 3.8 | 6.6 | 9.77 |
| OSFASW_M | 5.2 | 5.2 | 7 | 9 | 6.6 | 12.2 | 7.53 |

TABLE IV. THE AVERAGE CLASSIFICATION ACCURACY (%) ON $\theta=0.1$

| Dataset | OS²FSU | LOSSA | SFS_FI_Z | OSFS_Z | SAOLA_Z | OSFASW_Z | SFS_FI_M | OSFS_M | SAOLA_M | OSFASW_M |
|---|---|---|---|---|---|---|---|---|---|---|
| D1 | **90.05±0.56** | 83.95±0.89 | 66.02±0.59 | 85.32±0.82 | 79.12±0.73 | 83.92±1.93 | 59.57±1.49 | 84.03±0.54 | 67.95±0.77 | 70.13±0.91 |
| D2 | **72.78±1.57** | 69.84±1.01 | 56.72±4.69 | 54.15±3.27 | 63.26±3.46 | 58.7±2.14 | 69.15±2.13 | 64.87±1.00 | 71.44±2.56 | 65.16±1.14 |
| D3 | **89.64±1.22** | 86.48±2.69 | 87.01±3.49 | 82.9±1.97 | 88.12±1.83 | 84.39±1.3 | 88.47±2.45 | 86.47±1.67 | 87.49±1.81 | 84.54±1.24 |
| D4 | **86.26±2.48** | 77.44±4.66 | 80.56±3.59 | 81.13±4.05 | 57.93±7.07 | 78.72±4.67 | 77.1±3.64 | 77.33±4.44 | 58±6.12 | 72.15±4.94 |
| D5 | **82.8±8.01** | 71.97±11.3 | 63.45±6.81 | 75.56±12.5 | 66.49±11.0 | 76.19±12.2 | 62.47±8.62 | 72.2±11.3 | 57.02±5.75 | 71.42±11.5 |
| D6 | **73.25±3.19** | 50.66±4.51 | 34.53±2.77 | 54±6.91 | 36±3.44 | 47.56±6.38 | 34.74±2.08 | 50.75±4.04 | 34±3.27 | 42.91±4.19 |
| Average | **82.46** | 73.39 | 64.71 | 72.17 | 65.15 | 71.58 | 65.25 | 72.61 | 62.65 | 67.72 |

TABLE V. THE AVERAGE CLASSIFICATION ACCURACY (%) ON D1

| θ | OS²FSU | LOSSA | SFS_FI_Z | OSFS_Z | SAOLA_Z | OSFASW_Z | SFS_FI_M | OSFS_M | SAOLA_M | OSFASW_M |
|---|---|---|---|---|---|---|---|---|---|---|
| 0.1 | **90.05±0.56** | 83.95±0.89 | 66.02±0.59 | 85.32±0.82 | 79.12±0.73 | 83.92±1.93 | 59.57±1.49 | 84.03±0.54 | 67.95±0.77 | 70.13±0.91 |
| 0.2 | **91.88±0.59** | 84.85±1.24 | 67.62±0.87 | 85.27±2.71 | 74.68±1.40 | 81.42±3.45 | 68.23±1.40 | 83.62±1.31 | 67.2±1.32 | 76.62±1.90 |
| 0.3 | **90.78±0.49** | 79.55±1.66 | 73.03±3.02 | 83.27±3.83 | 76.73±4.23 | 83.05±3.88 | 66.77±1.57 | 80.92±1.33 | 66.22±1.85 | 79.55±1.09 |
| 0.4 | **90.75±1.16** | 82.05±1.16 | 69.52±4.66 | 77.07±8.20 | 71.82±8.48 | 77.52±8.03 | 59.48±2.23 | 83.03±1.12 | 67.55±2.31 | 81.45±1.07 |
| 0.5 | **90.22±1.00** | 84.68±1.18 | 66.27±9.34 | 70.77±9.20 | 72.53±11.3 | 70.38±9.51 | 69.95±2.10 | 84.87±0.33 | 62.97±2.43 | 82.45±0.77 |
| 0.6 | **87.95±1.07** | 83.32±1.05 | 65.28±11.1 | 64.72±12.6 | 67.12±12.9 | 65.12±12.5 | 65.87±2.81 | 85.07±0.15 | 74.9±1.52 | 82.87±0.55 |
| 0.7 | **87.22±1.93** | 80.52±0.80 | 54.7±14.0 | 56.25±14.3 | 55.63±17.6 | 56.72±14.2 | 67.78±3.34 | 83.18±0.87 | 72.47±1.62 | 81.82±0.51 |
| 0.8 | **77.47±3.47** | 69.95±2.40 | 39.68±17.3 | 41.6±16.3 | 41.63±21.0 | 39.48±15.5 | 58.78±3.30 | 72.33±7.56 | 70.78±4.63 | 72.33±7.22 |
| 0.9 | **61.05±9.87** | 48.88±4.19 | 28.13±15.1 | 27.23±13.1 | 29.1±15.9 | 27±13.2 | 49.85±13.9 | 54.63±17.0 | 54.33±15.9 | 54.77±17.3 |
| Average | **85.26** | 77.53 | 58.92 | 65.72 | 63.15 | 64.96 | 62.92 | 79.08 | 67.15 | 75.78 |

TABLE VI. THE RANK SUM OF THE WILCOXON SIGNED-RANKS*

| $\theta$ | LOSSA | | SFS_FI_Z | | OSFS_Z | | SAOLA_Z | | OSFASW_Z | | SFS_FI_M | | OSFS_M | | SAOLA_M | | OSFASW_M | |
|---|---|---|---|---|---|---|---|---|---|---|---|---|---|---|---|---|---|---|
| | $R^+$ | $R^-$ | $R^+$ | $R^-$ | $R^+$ | $R^-$ | $R^+$ | $R^-$ | $R^+$ | $R^-$ | $R^+$ | $R^-$ | $R^+$ | $R^-$ | $R^+$ | $R^-$ | $R^+$ | $R^-$ |
| 0.1 | 21 | 0 | 21 | 0 | 21 | 0 | 21 | 0 | 21 | 0 | 21 | 0 | 21 | 0 | 21 | 0 | 21 | 0 |
| 0.2 | 21 | 0 | 21 | 0 | 21 | 0 | 21 | 0 | 21 | 0 | 21 | 0 | 21 | 0 | 21 | 0 | 21 | 0 |
| 0.3 | 21 | 0 | 21 | 0 | 21 | 0 | 21 | 0 | 21 | 0 | 21 | 0 | 21 | 0 | 21 | 0 | 21 | 0 |
| 0.4 | 21 | 0 | 21 | 0 | 21 | 0 | 21 | 0 | 21 | 0 | 21 | 0 | 21 | 0 | 21 | 0 | 21 | 0 |
| 0.5 | 21 | 0 | 21 | 0 | 21 | 0 | 21 | 0 | 21 | 0 | 21 | 0 | 21 | 0 | 21 | 0 | 21 | 0 |
| 0.6 | 21 | 0 | 21 | 0 | 21 | 0 | 21 | 0 | 21 | 0 | 20 | 1 | 21 | 0 | 19 | 2 | 21 | 0 |
| 0.7 | 21 | 0 | 21 | 0 | 21 | 0 | 21 | 0 | 21 | 0 | 21 | 0 | 21 | 0 | 21 | 0 | 21 | 0 |
| 0.8 | 21 | 0 | 21 | 0 | 21 | 0 | 21 | 0 | 21 | 0 | 20 | 1 | 16 | 5† | 19 | 2 | 15 | 6† |
| 0.9 | 21 | 0 | 21 | 0 | 21 | 0 | 21 | 0 | 21 | 0 | 17 | 4† | 15 | 6† | 15 | 6† | 15 | 6† |

\* If the smaller of the $R^+$ and $R^-$ is greater than 2, the null-hypothesis will be accepted.
† The position of $R^-$ which is larger than 2.

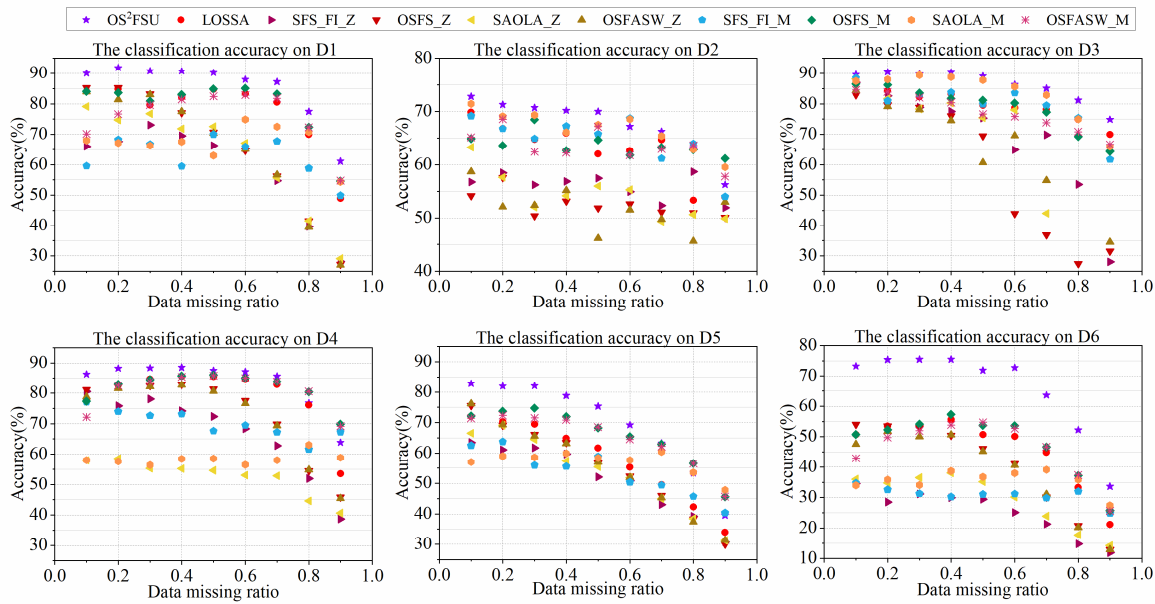

Fig. 3. The average accuracy comparison of three classifier on each dataset.

## VI. CONCLUSION

In this paper, we handle the uncertainty in OS²FS by incorporating fuzzy membership function and neighborhood rough set into online sparse streaming feature selection. Then, we propose an online sparse streaming feature selection with uncertainty (OS²FSU) algorithm. In OS²FSU, LFA is utilized to pre-estimate the missing data in sparse streaming features at first, then fuzzy logic and neighborhood rough set are employed to alleviate the uncertainty between estimated streaming features and labels during conducting feature selection. Compared with other state-of-the-art algorithms on six datasets, we achieve high prediction accuracy. Notably, when the missing data ratio is larger than 0.7, the performance of all the algorithms dramatically decrease. Next, we will research more advanced missing data estimation methods to improve OS²FS.

## REFERENCES

[1] S. Lall, D. Sinha, A. Ghosh, D. Sengupta, and S. Bandyopadhyay, "Stable feature selection using copula based mutual information," *Pattern Recognition*, vol. 112, p. 107697, 2021.

[2] X. Song, Y. Zhang, D. Gong, and X. Sun, "Feature selection using bare-bones particle swarm optimization with mutual information," *Pattern Recognition*, vol. 112, p. 107804, 2021.

[3] W. Ji et al., "Fuzzy rough sets and fuzzy rough neural networks for feature selection: A review," *Wiley Interdisciplinary Reviews: Data Mining and Knowledge Discovery*, vol. 11, no. 3, p. e1402, 2021.

[4] H. Zhao, P. Wang, Q. Hu, and P. Zhu, "Fuzzy Rough Set Based Feature Selection for Large-Scale Hierarchical Classification," *IEEE Trans. Fuzzy Syst.*, vol. 27, no. 10, pp. 1891–1903, Oct. 2019, doi: 10.1109/TFUZZ.2019.2892349.

[5] L. Sun, L. Wang, W. Ding, Y. Qian, and J. Xu, "Feature Selection Using Fuzzy Neighborhood Entropy-Based Uncertainty Measures for Fuzzy Neighborhood Multigranulation Rough Sets," *IEEE Trans. Fuzzy Syst.*, vol. 29, no. 1, pp. 19–33, Jan. 2021, doi: 10.1109/TFUZZ.2020.2989098.

[6] X. Yan and M. Jia, "Intelligent fault diagnosis of rotating machinery using improved multiscale dispersion entropy and mRMR feature selection," *Knowledge-Based Systems*, vol. 163, pp. 450–471, 2019.

[7] X. Wu, K. Yu, W. Ding, H. Wang, and X. Zhu, "Online Feature Selection with Streaming Features," *IEEE Transactions on Pattern Analysis and Machine Intelligence*, vol. 35, no. 5, pp. 1178–1192, 2013, doi: 10.1109/TPAMI.2012.197.

[8] K. Yu, X. Wu, W. Ding, and J. Pei, "Scalable and accurate online feature selection for big data," *ACM Transactions on Knowledge Discovery from Data (TKDD)*, vol. 11, no. 2, pp. 1–39, 2016.

[9] D. You et al., "Online Feature Selection for Streaming Features Using Self-Adaption Sliding-Window Sampling," *IEEE Access*, vol. 7, pp. 16088–16100, 2019, doi: 10.1109/ACCESS.2019.2894121.

[10] P. Zhou, P. Li, S. Zhao, and X. Wu, "Feature Interaction for Streaming Feature Selection," *IEEE Trans. Neural Netw. Learning Syst.*, vol. 32, no. 10, pp. 4691–4702, Oct. 2021, doi: 10.1109/TNNLS.2020.3025922.

[11] X. Li, Y. Wang, and R. Ruiz, "A Survey on Sparse Learning Models for Feature Selection," *IEEE Transactions on Cybernetics*, vol. 52, no. 3, pp. 1642–1660, 2022, doi: 10.1109/TCYB.2020.2982445.

[12] D. Wu, Y. He, X. Luo, and M. Zhou, "A Latent Factor Analysis-Based Approach to Online Sparse Streaming Feature Selection," *IEEE Trans. Syst. Man Cybern, Syst.*, pp. 1–15, 2021, doi: 10.1109/TSMC.2021.3096065.

[13] X. Luo, Y. Zhou, Z. Liu, L. Hu, and M. Zhou, "Generalized Nesterov's Acceleration-incorporated Non-negative and Adaptive Latent Factor Analysis," *IEEE Transactions on Services Computing*, pp. 1–1, 2021, doi: 10.1109/TSC.2021.3069108.

[14] X. Luo, Z. Wang, and M. Shang, "An instance-frequency-weighted regularization scheme for non-negative latent factor analysis on high-dimensional and sparse data," *IEEE Transactions on Systems, Man, and Cybernetics: Systems*, vol. 51, no. 6, pp. 3522–3532, 2019.

[15] X. Luo, Y. Zhou, Z. Liu, and M. Zhou, "Fast and Accurate Non-negative Latent Factor Analysis on High-dimensional and Sparse Matrices in Recommender Systems," *IEEE Transactions on Knowledge and Data Engineering*, pp. 1–1, 2021, doi: 10.1109/TKDE.2021.3125252.

[16] D. Wu and X. Luo, "Robust Latent Factor Analysis for Precise Representation of High-Dimensional and Sparse Data," *IEEE/CAA Journal of Automatica Sinica*, vol. 8, no. 4, pp. 796–805, 2021, doi: 10.1109/JAS.2020.1003533.

[17] X. Luo, H. Wu, H. Yuan, and M. Zhou, "Temporal Pattern-Aware QoS Prediction via Biased Non-Negative Latent Factorization of Tensors," *IEEE Transactions on Cybernetics*, vol. 50, no. 5, pp. 1798–1809, 2020, doi: 10.1109/TCYB.2019.2903736.

[18] H. Wu, X. Luo, and M. Zhou, "Advancing Non-Negative Latent Factorization of Tensors With Diversified Regularization Schemes," *IEEE Transactions on Services Computing*, vol. 15, no. 3, pp. 1334–1344, 2022, doi: 10.1109/TSC.2020.2988760.

[19] D. Wu, Q. He, X. Luo, M. Shang, Y. He, and G. Wang, "A Posterior-Neighborhood-Regularized Latent Factor Model for Highly Accurate Web Service QoS Prediction," *IEEE Transactions on Services Computing*, vol. 15, no. 2, pp. 793–805, 2022, doi: 10.1109/TSC.2019.2961895.

[20] D. Wu, X. Luo, M. Shang, Y. He, G. Wang, and X. Wu, "A Data-Characteristic-Aware Latent Factor Model for Web Services QoS Prediction," *IEEE Transactions on Knowledge and Data Engineering*, vol. 34, no. 6, pp. 2525–2538, 2022, doi: 10.1109/TKDE.2020.3014302.

[21] X. Luo, H. Wu, Z. Wang, J. Wang, and D. Meng, "A Novel Approach to Large-Scale Dynamically Weighted Directed Network Representation," *IEEE Transactions on Pattern Analysis and Machine Intelligence*, pp. 1–1, 2021, doi: 10.1109/TPAMI.2021.3132503.

[22] X. Luo, M. Zhou, S. Li, L. Hu, and M. Shang, "Non-Negativity Constrained Missing Data Estimation for High-Dimensional and Sparse Matrices from Industrial Applications," *IEEE Transactions on Cybernetics*, vol. 50, no. 5, pp. 1844–1855, 2020, doi: 10.1109/TCYB.2019.2894283.

[23] X. Luo, M. Zhou, Z. Wang, Y. Xia, and Q. Zhu, "An Effective Scheme for QoS Estimation via Alternating Direction Method-Based Matrix Factorization," *IEEE Transactions on Services Computing*, vol. 12, no. 4, pp. 503–518, 2019, doi: 10.1109/TSC.2016.2597829.

[24] P. Zhou, X. Hu, P. Li, and X. Wu, "OFS-Density: A novel online streaming feature selection method," *Pattern Recognition*, vol. 86, pp. 48–61, Feb. 2019, doi: 10.1016/j.patcog.2018.08.009.

[25] P. Zhou, X. Hu, P. Li, and X. Wu, "Online streaming feature selection using adapted Neighborhood Rough Set," *Information Sciences*, vol. 481, pp. 258–279, May 2019, doi: 10.1016/j.ins.2018.12.074.

[26] X. Luo, Y. Yuan, S. Chen, N. Zeng, and Z. Wang, "Position-Transitional Particle Swarm Optimization-Incorporated Latent Factor Analysis," *IEEE Transactions on Knowledge and Data Engineering*, vol. 34, no. 8, pp. 3958–3970, 2022, doi: 10.1109/TKDE.2020.3033324.

[27] X. Luo, M. Shang, and S. Li, "Efficient Extraction of Non-negative Latent Factors from High-Dimensional and Sparse Matrices in Industrial Applications," in *2016 IEEE 16th International Conference on Data Mining (ICDM)*, 2016, pp. 311–319. doi: 10.1109/ICDM.2016.0042.

[28] D. Wu, X. Luo, M. Shang, Y. He, G. Wang, and M. Zhou, "A Deep Latent Factor Model for High-Dimensional and Sparse Matrices in Recommender Systems," *IEEE Transactions on Systems, Man, and Cybernetics: Systems*, vol. 51, no. 7, pp. 4285–4296, 2021, doi: 10.1109/TSMC.2019.2931393.

[29] D. Wu, P. Zhang, Y. He, and X. Luo, "A Double-Space and Double-Norm Ensembled Latent Factor Model for Highly Accurate Web Service QoS Prediction," *IEEE Transactions on Services Computing*, pp. 1–1, 2022, doi: 10.1109/TSC.2022.3178543.

[30] D. Wu, M. Shang, X. Luo, and Z. Wang, "An $L_1$-and-$L_2$-Norm-Oriented Latent Factor Model for Recommender Systems," *IEEE Trans. Neural Netw. Learning Syst.*, pp. 1–14, 2021, doi: 10.1109/TNNLS.2021.3071392.

[31] L. Hu, X. Yuan, X. Liu, S. Xiong, and X. Luo, "Efficiently Detecting Protein Complexes from Protein Interaction Networks via Alternating Direction Method of Multipliers," *IEEE/ACM Transactions on Computational Biology and Bioinformatics*, vol. 16, no. 6, pp. 1922–1935, 2019, doi: 10.1109/TCBB.2018.2844256.

[32] L. Xin, Y. Yuan, M. Zhou, Z. Liu, and M. Shang, "Non-Negative Latent Factor Model Based on β-Divergence for Recommender Systems," *IEEE Transactions on Systems, Man, and Cybernetics: Systems*, vol. 51, no. 8, pp. 4612–4623, 2021, doi: 10.1109/TSMC.2019.2931468.

[33] X. Luo, Z. Liu, M. Shang, J. Lou, and M. Zhou, "Highly-Accurate Community Detection via Pointwise Mutual Information-Incorporated Symmetric Non-Negative Matrix Factorization," *IEEE Transactions on Network Science and Engineering*, vol. 8, no. 1, pp. 463–476, 2021, doi: 10.1109/TNSE.2020.3040407.

[34] X. Shi, Q. He, X. Luo, Y. Bai, and M. Shang, "Large-Scale and Scalable Latent Factor Analysis via Distributed Alternative Stochastic Gradient Descent for Recommender Systems," *IEEE Transactions on Big Data*, vol. 8, no. 2, pp. 420–431, 2022, doi: 10.1109/TBDATA.2020.2973141.

[35] Y. Yuan, Q. He, X. Luo, and M. Shang, "A Multilayered-and-Randomized Latent Factor Model for High-Dimensional and Sparse Matrices," *IEEE Transactions on Big Data*, vol. 8, no. 3, pp. 784–794, 2022, doi: 10.1109/TBDATA.2020.2988778.

[36] Z. Liu, X. Luo, and Z. Wang, "Convergence Analysis of Single Latent Factor-Dependent, Nonnegative, and Multiplicative Update-Based Nonnegative

Latent Factor Models," *IEEE Transactions on Neural Networks and Learning Systems*, vol. 32, no. 4, pp. 1737–1749, 2021, doi: 10.1109/TNNLS.2020.2990990.

[37] X. Luo, H. Wu, and Z. Li, "NeuLFT: A Novel Approach to Nonlinear Canonical Polyadic Decomposition on High-Dimensional Incomplete Tensors," *IEEE Transactions on Knowledge and Data Engineering*, pp. 1–1, 2022, doi: 10.1109/TKDE.2022.3176466.

[38] X. Luo, J. Sun, Z. Wang, S. Li, and M. Shang, "Symmetric and nonnegative latent factor models for undirected, high-dimensional, and sparse networks in industrial applications," *IEEE Transactions on Industrial Informatics*, vol. 13, no. 6, pp. 3098–3107, 2017.

[39] J. Wu, L. Chen, Y. Feng, Z. Zheng, M. C. Zhou, and Z. Wu, "Predicting quality of service for selection by neighborhood-based collaborative filtering," *IEEE Transactions on Systems, Man, and Cybernetics: Systems*, vol. 43, no. 2, pp. 428–439, 2012.

[40] H. Wu, Z. Zhang, K. Yue, B. Zhang, J. He, and L. Sun, "Dual-regularized matrix factorization with deep neural networks for recommender systems," *Knowledge-Based Systems*, vol. 145, pp. 46–58, 2018.

[41] M. Shang, Y. Yuan, X. Luo, and M. Zhou, "An α–β-Divergence-Generalized Recommender for Highly Accurate Predictions of Missing User Preferences," *IEEE Transactions on Cybernetics*, vol. 52, no. 8, pp. 8006–8018, 2022, doi: 10.1109/TCYB.2020.3026425.

[42] K. B. Narayanan and S. Muthusamy, "Prediction of machinability parameters in turning operation using interval type-2 fuzzy logic system based on semi-elliptic and trapezoidal membership functions," *Soft Computing*, pp. 1–20, 2022.

[43] A. Jain and A. Sharma, "Membership function formulation methods for fuzzy logic systems: A comprehensive review," *Journal of Critical Reviews*, vol. 7, no. 19, pp. 8717–8733, 2020.

[44] J. Serrano-Guerrero, F. P. Romero, and J. A. Olivas, "Fuzzy logic applied to opinion mining: a review," *Knowledge-Based Systems*, vol. 222, p. 107018, 2021.

[45] J. X. Deng and Y. Deng, "Information Volume of Fuzzy Membership Function.," *International Journal of Computers, Communications & Control*, vol. 16, no. 1, 2021.

[46] J. Demšar, "Statistical comparisons of classifiers over multiple data sets," *The Journal of Machine Learning Research*, vol. 7, pp. 1–30, 2006.